\documentclass{article} 
\usepackage{iclr2026_conference,times}
\usepackage{graphicx}
\usepackage{float}

\usepackage{amsmath,amsfonts,bm}

\def\eqref#1{equation~\ref{#1}}

\def\1{\bm{1}}

\DeclareMathAlphabet{\mathsfit}{\encodingdefault}{\sfdefault}{m}{sl}
\SetMathAlphabet{\mathsfit}{bold}{\encodingdefault}{\sfdefault}{bx}{n}

\usepackage{hyperref}
\usepackage{url}
\usepackage{algorithm}
\usepackage{algpseudocode}
\usepackage{multirow}
\usepackage{booktabs}
\usepackage{bbding}
\usepackage{float}
\usepackage{subcaption}

\usepackage{colortbl}

\title{Spatial-Spectral Binarized Neural Network for Panchromatic and Multi-spectral Images Fusion}

\author{Yizhen Jiang$^1$, Mengting Ma$^1$ \thanks{Email : 
mtma@zju.edu.cn}, Anqi Zhu$^1$, Xiaowen Ma$^1$, Jiaxin Li$^2$, Wei Zhang$^1$ \\
$^1$ Zhejiang University, China\\
$^2$ Chongqing University of Posts and Telecommunications, China\\
}

\begin{document}

\maketitle

\begin{abstract}
Remote sensing pansharpening aims to reconstruct spatial-spectral properties during the fusion of panchromatic (PAN) images and low-resolution multi-spectral (LR-MS) images, finally generating the high-resolution multi-spectral (HR-MS) images. Although deep learning-based models have achieved excellent performance, they often come with high computational complexity, which hinder their applications on resource-limited devices. In this paper, we explore the feasibility of applying the binary neural network (BNN) to pan-sharpening. Nevertheless, there are two main issues with binarizing pan-sharpening models: (\emph{i}) the binarization will cause serious spectral distortion due to the inconsistent spectral distribution of the PAN/LR-MS images; (\emph{ii}) the common binary convolution kernel is difficult to adapt to the multi-scale and anisotropic spatial features of remote sensing objects, resulting in serious degradation of contours.
To address the above issues, we design the customized spatial-spectral binarized convolution (S2B-Conv), which is composed of the Spectral-Redistribution Mechanism (SRM) and Gabor Spatial Feature Amplifier (GSFA). Specifically, SRM employs an affine transformation, generating its scaling and bias parameters through a dynamic learning process. GSFA, which randomly selects different frequencies and angles within a preset range, enables to better handle multi-scale and-directional spatial features.
A series of S2B-Conv form a brand-new binary network for pan-sharpening, dubbed as S2BNet. Extensive quantitative and qualitative experiments have shown our high-efficiency binarized pan-sharpening method can attain a promising performance. The code link is \textcolor{blue}{https://github.com/Ritayiyi/S2BNet}.

\end{abstract}

\section{Introduction}

With the increasing demand of earth observation and monitoring, existing optical satellites (e.g., GaoFen-2, QuickBird) can simultaneously record bundled low-resolution multispectral (LR-MS) and high-resolution panchromatic (PAN) images from the same scene. Due to physical limitations of existing satellite sensors, the recorded LR-MS image usually includes rich spectral property but relatively sparse spatial property, while their corresponding PAN image contains abundant spatial property but sparse spectral property. Since the remote sensing image with rich spatial-spectral properties, i.e., high-resolution multispectral (HR-MS) image, are crucial for practical applications~\cite{li2022deep,han2024gretnet,Asokan_Anitha_2019}, pansharpening technique, which could obtain the HR-MS image by reconstructing spatial-spectral properties during the fusion of the recorded PAN and LR-MS images, has been widely explored~\cite{Xing_Zhang_He_Zhang_Zhang_2023}. 

Existing state-of-the-art (SOTA) pan-sharpening methods are based on deep learning. Convolutional neural network (CNN) and Transformer~\cite{Wu_Huang_Deng_Zhang_2022, tan2024revisiting,DBLP:conf/ijcai/LiLXHY23} have been employed as powerful models to reconstruct spatial-spectral properties for target HR-MS images. For example, SRPPNN~\cite{cai2020super} performs the pan-sharpening learning using a deeper CNN network architecture implemented residual learning. Zhou~\textit{et al}. propose a series of Transformer-based algorithms push the performance boundary again~\cite{zhou2022pan}. Although superior performance is achieved, these CNN-/Transformer-based methods require powerful hardwares with abundant computing and memory resources, such as costing 439.38M in HyperTransformer~\cite{bandara2022hypertransformer}. This motivates us to reduce the memory and computational burden of pan-sharpening methods while preserving the performance as much as possible so that the algorithms can be deployed on resource-limited devices.

To improve the efficiency of deep neural networks, many network compression techniques are proposed, including network quantization~\cite{qin2020binary}, parameter pruning~\cite{wang2022trainability} and knowledge distillation~\cite{wang2022makes}. Among these approaches, Binarized neural network (BNN) stands out as an extreme case of network quantization, which binarizes both weights and activations into only $1$-bit. In particular, the foundation of BNNs lies in their pure logical computations, primarily XNOR and bit-count operations. This makes them highly energy-efficient, particularly for embedded devices~\cite{zhang2024binarized}. However, directly applying model binarization for pan-sharpening algorithms may face several challenges.~\textbf{(i)} The density and distribution of the spectra within the LR-MS image are different, similarly, the spectral density and distribution between LR-MS and PAN images are also different. During the fusion of LR-MS and PAN images, equally binarizing the activations of different spectral channels may lead to severe distortion of spectral features.~\textbf{(ii)} Remote sensing scenes are inherently multi-scale and anisotropic. Directly using the common binary convolutions, such as those used in BNNs, has two drawbacks: \emph{{(1})} the receptive field is limited, making it difficult to understand the global structure, resulting in blurred or disjointed contours; and \emph{(2)} the convolution kernel is isotropic, which means it learns an "averaged" feature response during training, which can easily produce jagged effects or blurred edges.

Bearing the above challenges in mind, we propose a novel binarized method, namely Spatial-Spectral Binarized Neural Network (S2BNet) for efficient remote sensing pan-sharpening. Different from the recent binarized multi-spectral fusion network~\cite{hou2025binarized}, our work does not include complex computations like diffusion operation and attention mechanism that are difficult to implement on edge devices. \textbf{Firstly}, we design the basic unit, spatial-spectral binarized convolution (S2B-Conv), used in model binarization. S2B-Conv can adapt inconsistent spectral distribution and diverse spatial features before binarizing the activation. \textbf{Secondly}, the Spectral-Redistribution Mechanism (SRM) in S2B-Conv is employed to generate scaling and bias parameters in a adaptive learning manner, adjusting the density and distribution of spectral bands. \textbf{Thirdly}, we propose the Gabor Spatial Feature Amplifier (GSFA) in S2B-Conv, which captures multi-scale and-directional spatial features by randomly selecting different frequencies and angles within a preset range. \textbf{Finally}, we derive our S2BNet by using the proposed spatial-spectral binarized convolution to binarize the base model. As shown in Tab.~\ref{tab:gf2}, S2BNet outperforms SOTA BNNs by large margins, over $2.49$ dB.
In a nutshell, our contributions can be summarized as follows:
\begin{itemize}
    \item We propose a novel BNN-based algorithm S2BNet, which is composed of several S2B-Conv units, for remote sensing pan-sharpening. In particular, the S2B-Conv could adapt complex spatial-spectral features before binarizing the model.

    \item The Spectral-Redistribution Mechanism (SRM) in S2B-Conv is introduced to adjust the distribution in spectral dimension through adaptive learning process.
    
    \item The Gabor Spatial Feature Amplifier (GSFA) in S2B-Conv is introduced to learn multi-scale and multi-directional spatial features by randomly selecting different frequencies and angles.
  
    \item Experiments on multiple remote sensing benchmark datasets show that the proposed S2BNet outperforms state-of-the-art binary neural networks.
\end{itemize}

\section{Related Work}
\subsection{Pan-sharpening}
The classic pan-sharpening methods mainly consist of $3$ categories: CS-based~\cite{Gillespie_Kahle_Walker_1987}, MRA-based~\cite{Nunez_Otazu_Fors_Prades_Pala_Arbiol_1999}, and VO-based methods~\cite{Fasbender_Radoux_Bogaert_2008}. Deep learning-based methods have significantly improved the performance of pan-sharpening tasks in recent times. As a groundbreaking study, PNN~\cite{Masi_Cozzolino_Verdoliva_Scarpa_2016} first introduced CNN-based methods into pan-sharpening tasks. PANNet~\cite{Yang_Fu_Hu_Huang_Ding_Paisley_2017} leverages residual connections and high-frequency filtering techniques within a CNN framework. Moreover, HyperTransformer~\cite{DBLP:conf/cvpr/BandaraP22} was one of the early attempts to introduce Transformer-based methods into this field. PanFormer~\cite{zhou2022panformer} utilizes the Transformer framework to 
improve spatial resolution while maintaining spectral integrity. 

Although deep learning models have achieved remarkable progress in pan-sharpening over traditional methods, they typically require powerful hardware with considerable memory and computational capacity, that hinders their deployment on resource-limited satellites. In this study, we explore the potential of using binarized neural networks for pan-sharpening to make the model more lightweight.

\subsection{Binarized Neural Network}
Binary Neural Networks (BNNs) have emerged as a promising solution for deploying deep learning models on resource-limited devices~\cite{zhang2024binarized}. By leveraging binary representations for weights and activations, BNNs achieve substantial reductions in memory footprint and computational requirements~\cite{hubara2016binarized}.

The pioneering method~\cite{Hubara2016BNN} employs the sign function to derive binarized activations and weights, and it utilizes the straight-through estimator~\cite{bengio2013estimating} for optimizing network parameters. Nevertheless, BNNs encounter precision loss attributable to quantization error and gradient error~\cite{zhang2024binarized}. To tackle the quantization error, XNOR-Net~\cite{rastegari2016xnor} incorporates scaling factors for both weights and activations. To mitigate the gradient error, Liu et al.~\cite{liu2018bi} propose employing a piecewise quadratic function. 

Recently, exploration of binary networks for low-level vision applications has drawn great attention. Xin et al.~\cite{xin2020binarized} propose a novel binarization method with a bit accumulation mechanism to approximate full-precision convolution for image super-resolution task. Work~\cite{zhang2024binarized} investigates the use of BNNs for low-light raw video enhancement, facing the challenges of efficiently fusing temporal information and bridging the performance gap with full-precision models.

The inherent resource constraints of remote sensing satellites make model compression a necessity. However, there is hardly any research devoted to light-weighting of remote sensing. 

\begin{figure}[t]
    \centering
    \includegraphics[width=\linewidth]{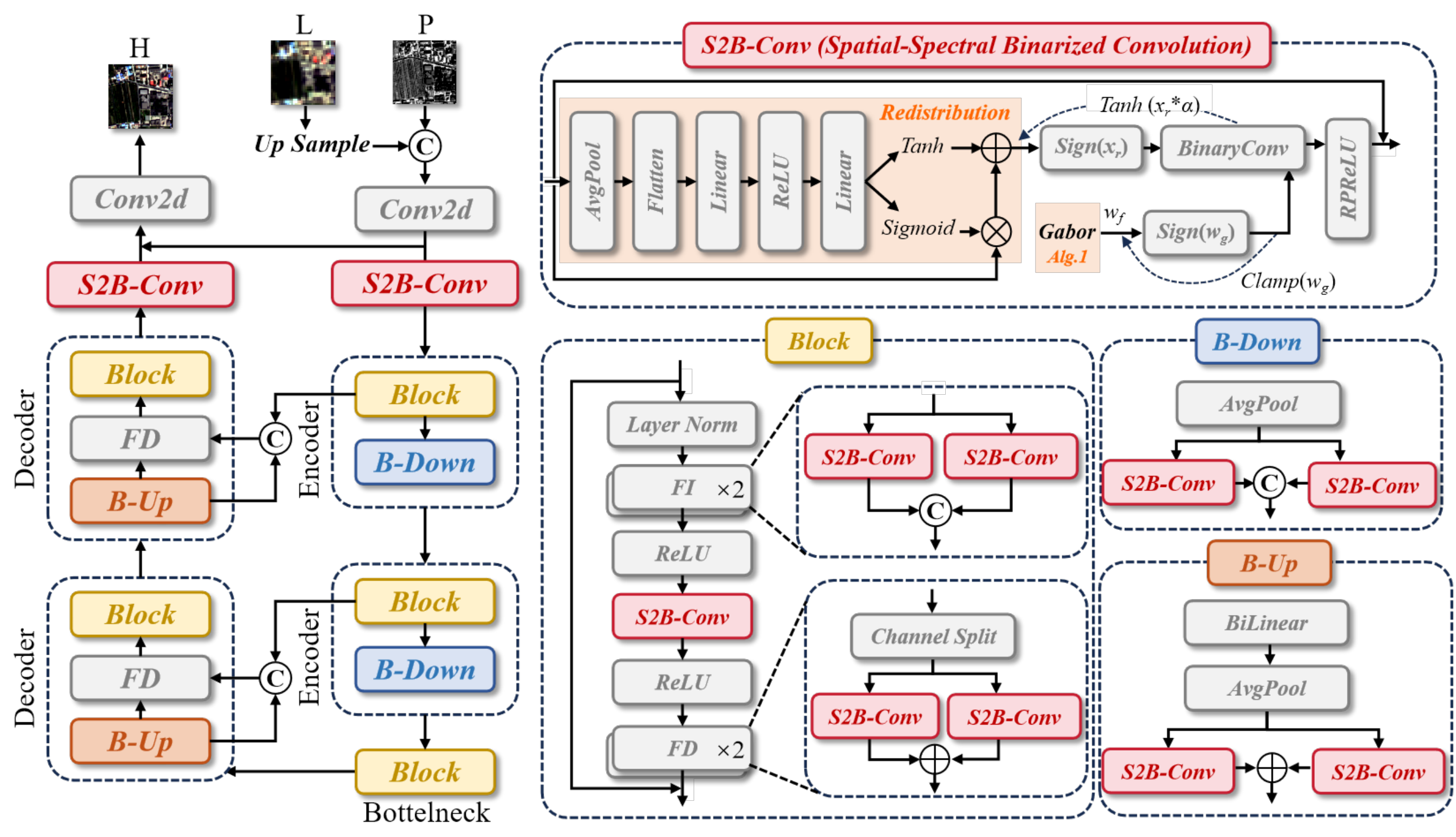}
    \caption{The overall frame of our method. Our model primarily consists of the novel convolution (S2B-Conv), which incorporates two core modules: the Spectral-Redistribution Mechanism (denoted as "Redistribution") and the Gabor Spatial Feature Amplifier (denoted as "Gabor"). The specific implementation of Gabor is detailed in Algorithm 1.}
    \label{fig:framework}
\end{figure}

\section{Method}
In this section, we present the overall architecture of our network for pan-sharpening. Then, we elaborate on the details of the spatial-spectral binarized convolution, which mainly consists of the Spectral-Redistribution Mechanism and Gabor Spatial Feature Amplifier.

\subsection{Overall Architecture}
Binary networks have achieved significant success in the fields of hyperspectral image reconstruction and low-light raw video enhancement. However, few studies have attempted to apply them to the task of pan-sharpening. Building on the existing research in low-level vision tasks and considering the characteristics of pan-sharpening, we design a network that focuses on both spatial and spectral attributes, as the reconstruction quality of these attributes greatly affects the outcome of pan-sharpening.

The general architecture of our S2BNet network is shown in Fig.~\ref{fig:framework}. Inspired by~\cite{cai2023binarized}, we use a U-shaped structure for the base model. First, we upsample the LR-MS images and concatenate them with the Pan images, and feed the output into the full-precision convolutional layer to generate a shallow feature $X_s \in \mathbb{R}^{H \times W \times C}$ and subsequently into the S2B-Conv block. Here, $H$, $W$, and $C$ denote the height, width, and the number of channels. Then it undergoes two encoders, a bottleneck, and two decoders. To alleviate the information loss during processing, skip connections between the encoders and the decoders are employed. After that, the S2B-Conv block maps the features to generate the deep feature $X_d \in \mathbb{R}^{H \times W \times C} $. Finally, the sum of $X_s$ and $X_d$ is fed into another full-precision convolutional layer to produce the reconstructed HR-MS images. Specifically, each encoder consists of a basic block followed by a B-Down block, while each decoder is composed of a B-Up block, a Fusion Decrease block, and a basic block. Furthermore, the basic block serves as the bottleneck. It is worth noting that the entire network only employs two full-precision convolutions, while all other convolutions are binary convolutions.

As illustrated in Fig.~\ref{fig:framework}, the Fusion Increase and Fusion Decrease modules both use binarized convolutional layers with a kernel size of 1 to aggregate the feature maps and modify the channels. The upsample module employs bilinear interpolation followed by binarized convolutional layers with a kernel size of 3 to upscale the feature maps and halve the channels, while the downsample module uses binarized convolutional layers with a kernel size of 3 to downscale the feature maps and double the channels. Moreover, the structure of the basic block is also presented in Fig.~\ref{fig:framework}. It consists of Layer Norm, Fusion Increase block, S2B-Conv block, Fusion Decrease block, and ReLU activation function. The S2B-Conv block will be thoroughly discussed in subsequent chapters.

\subsection{spatial-spectral binarized convolution}
As shown in the Fig.~\ref{fig:framework}, we first adjust the spectral distribution of the full-precision activations $X_f \in \mathbb{R}^{H \times W \times C}$ using the Spectral-Redistribution Mechanism to get $X_r \in \mathbb{R}^{H \times W \times C}$ (which will be introduced in Section~\ref{sec:redistribution}), and generate weights $W_f$ using the Gabor Spatial Feature Amplifier (which will be introduced in Section~\ref{sec:gabor}). Then, referring to~\cite{zhang2024binarized, cai2023binarized}, we binarize both $X_r$ and $W_f$, and fuse these two features using an XOR operation, as detailed subsequently.

The activation $X_r$ is binarized by a sign function in 1-bit values, producing $X_b \in \mathbb{R}^{H \times W \times C}$, where each element has a value of +1 or -1. The binarization procedure can be formulated as:
\begin{equation}
    X_b = \text{Sign}(X_r) = 
    \begin{cases} 
    +1, & X_r > 0 \\
    -1, & X_r \leq 0. 
    \end{cases}
\end{equation}
Since the Sign function is non-differentiable, we use a scalable hyperbolic tangent function to approximate the Sign function in the backpropagation as:
\begin{equation}
    x_b = \text{Tanh}(\alpha x_r) = \frac{e^{\alpha x_r} - e^{-\alpha x_r}}{e^{\alpha x_r} + e^{-\alpha x_r}},
\end{equation}
where $\alpha \in \mathbb{R}^+$is a learnable parameter that adaptively adjusts the distance between Tanh and Sign.

For weight $W_f$, we also utilize the sign function to generate the binarized weight $W_b$. And we adopt a piecewise linear function, Clip, during the backpropagation as:
\begin{equation}
   W_b= \text{Clip}(W_f) = 
    \begin{cases} 
    +1, &  W_f \geq 1 \\
    x, &  -1 < W_f < 1 \\
    -1, &  W_f \leq -1 
    \end{cases}
\end{equation}

Subsequently, the computationally intensive floating-point matrix multiplication operations in full-precision convolution can be substituted with pure logical XNOR and bit-count operations as:
\begin{equation}
    X_b \otimes W_b = \text{bitcount}(\text{XNOR}(X_b, W_b)).
\end{equation}

Following~\cite{zhang2024binarized}, we multiply the mean absolute value of the 32-bit weight value $W_f$ to narrow down the difference between binarized and full-precision weights, i.e.,
\begin{equation}
    S_i = \frac{\|W_f^i\|_1}{C \times K \times K}, \quad i = 1 \ldots C_{\text{out}},
\end{equation}
\begin{equation}
    Y = (X_b \otimes W_b) \odot S,
\end{equation}
where $S \in \mathbb{R}^{C_{\text{out}}}$ is the scaling factor and $Y \in \mathbb{R}^{H \times W \times C}$ denotes the output activation rescaled with the factors from weights.

To mitigate the loss of information from binarization, we introduce a residual connection that sums $X_f$ and $Y$. However, due to the significant difference in their value ranges, combining them directly using an identity mapping might obscure the information in $Y$. To resolve this issue, we first apply the $Y$ to RPReLU~\cite{liu2018bi}, to adjust its value range before the summation. This step can be represented as:
\begin{equation}
    X_o = X_f + \text{RPReLU}(Y),
\end{equation}
where $X_o \in \mathbb{R}^{H \times W \times C}$ is the output feature and RPReLU is formulated as:
\begin{equation}
\text{RPReLU}(Y) = \left\{
\begin{array}{ll}
Y_i - \gamma_i + \zeta_i, & Y_i > \zeta_i \\
\beta_i(Y_i - \gamma_i) + \zeta_i, & Y_i \leq \zeta_i,
\end{array}
\right.
\end{equation}
where $\gamma_i, \zeta_i, \beta_i \in \mathbb{R}^{C_{\text{out}}}$ are learnable parameters.

\subsubsection{The Spectral-Redistribution Mechanism}
\label{sec:redistribution}
Upon analyzing the LR-MS images, we observe that the distribution and density of spectral information differ significantly across various bands~\cite{cai2023binarized}. The spectral information of Pan images and LR-MS images also has significant differences. This inconsistency in spectral information can negatively impact the reconstruction of spectral features, leading to potential inaccuracies in the final output.

Existing studies attempt to adjust the distribution of spectral information using affine transformations. However, they predominantly utilize a randomly initialized approach for scaling and biasing, which does not adapt to the input data. These methods lack the ability to dynamically adjust parameters based on input variations, potentially leading to suboptimal performance across diverse datasets.

To address this limitation, we introduce a Spectral-Redistribution Mechanism that generates scaling and bias parameters through a data-driven mechanism. This approach not only enhances adaptability to varying input characteristics but also enhances feature reconstruction accuracy by optimizing the redistribution process through learning global features.

As depicted in Fig.~\ref{fig:framework}, first, the input feature $X_f$ undergoes global average pooling and is then flattened to obtain the global feature $X_g \in \mathbb{R}^{H \times W \times C}$. Subsequently, two fully connected layers along with a ReLU activation function are employed to generate the scaling factor $k$ and the bias $b$. Specifically, the number of channels of feature $X_g$ is first mapped from $C$ to $C/r$, and after passing through the ReLU, it is then mapped from $C/r$ to $2C$. Following this, the resulting feature is split along the channel dimension into two separate features, each with $C$ channels. Here $C$ and $r$ denote the number of channels and scaling factor, respectively. After that, we employ the sigmoid activation function to constrain the scaling factor $k$ within the range of 0 to 1, obtaining $k'$, and utilize the tanh function to limit the bias $b$ between -1 and 1, resulting in $b'$. Ultimately, the redistributed features are acquired through an affine transformation. The specific formulas are as follows:
\begin{equation}
        k' = \sigma(k) = \frac{1}{1 + e^{-k}},
\end{equation}
\begin{equation}
        b' = \tanh(b) = \frac{e^{b} - e^{-b}}{e^{b} + e^{-b}},
\end{equation}
\begin{equation}
        X_r = X_f \cdot k' + b'.
\end{equation}

\subsubsection{Gabor Spatial Feature Amplifier}
\label{sec:gabor}
The spectral characteristics of parking lots and open spaces, as well as bare land and leveled construction sites, are very similar and need to be distinguished by spatial features. However, their spatial shapes and sizes are also similar, and the resolution of the LR-MS images is too low. Therefore, a method that can extract spatial textures more accurately is needed.

To address this challenge, we propose the Gabor Spatial Feature Amplifier, which initializes the weights of convolutional layers using Gabor kernels. Gabor kernels, by randomly selecting different frequencies and angles within a preset range, enable subsequent convolutional layers to better handle multi-scale and multi-directional features, which is conducive to the learning of spatial features.

Although Gabor kernels are widely used in image processing to simulate the perception of edges and textures by the human visual system, existing research has not yet explored the application of Gabor kernels in binary convolutional networks. 

First, we define two lists: one containing different frequency values and another containing different angle values. For each output channel, we randomly select a frequency and an angle from these lists to generate the Gabor kernel. The Gabor kernel is generated based on the following formula:
\begin{equation}
        g(x, y; \lambda, \theta, \psi, \sigma, \gamma) = \exp\left(-\frac{x'^2 + \gamma^2 y'^2}{2\sigma^2}\right) \exp\left(i \left(2\pi \frac{x'}{\lambda} + \psi\right)\right),
\end{equation}
where
\begin{equation}
    \begin{cases}
    x' = x \cos \theta + y \sin \theta \\
    y' = -x \sin \theta + y \cos \theta.
    \end{cases}
\end{equation}
However, here we only use the real part, that is,
\begin{equation}
g(x, y; \lambda, \theta, \psi, \sigma, \gamma) = \exp\left(-\frac{x'^2 + \gamma^2 y'^2}{2\sigma^2}\right) \cos\left(2\pi \frac{x'}{\lambda} + \psi\right).
\end{equation}
The parameters of the Gabor function each play a distinct role in shaping the characteristics of the filter. The parameter $\lambda$ signifies the wavelength of the sinusoidal component within the Gabor function. $\theta$ indicates the orientation or direction of the filter. The phase offset of the sinusoidal function is denoted by $\psi$. The parameter $\sigma$ represents the standard deviation of the Gaussian envelope, and $\gamma$ is the aspect ratio that influences the ellipticity of the Gaussian function. Since each output channel selects different parameters, the $C_{out}$ output channels will have $C_{out}$ different Gabor kernels, thus enabling the learning of spatial features at different directions and scales. Because the generated Gabor kernels are intended to replace the weights in the convolutional layer, we need to normalize them so that their mean and variance are similar to the default weights of the convolutional layer. This ensures that the Gabor kernels are compatible with the initialization scheme typically used in convolutional neural networks. For the detailed algorithm, please refer to Algorithm.~\ref{alg:1}.
\begin{algorithm}
\caption{Gabor Spatial Feature Amplifier}
\label{alg:1}
\begin{algorithmic}[1]
\Require Number of input channels $in\_chn$, number of output channels $out\_chn$, number of groups $groups$, kernel size $kernel\_size$
\Ensure Gabor kernel $weight$
\While{not converged}
    \State $n\_in \gets in\_chn \times kernel\_size \times kernel\_size$ 
    \State Initialize frequency array $freqs$ with $4$ elements 
    \State Initialize angle array $thetas$ with $16$ elements
    \State $weight \gets \text{empty}(out\_chn, \frac{in\_chn}{groups}, kernel\_size, kernel\_size, \text{dtype=np.float32})$
    
    \For{$oc \gets 1$ to $out\_chn$} \Comment{For each output channel, we generate a kernel.}
        \State $\lambda \gets \text{random choice from } freqs$ 
        \State $\theta \gets \text{random choice from } thetas$ 
        \State $\gamma \gets \text{Set fixed parameter}$
        \State $\sigma \gets \text{Set fixed parameter based on } n_{in}$         
        \For{$x \gets 0$ to $kernel\_size - 1$}
            \For{$y \gets 0$ to $kernel\_size - 1$}
                \State $x' \gets x \cos(\theta) + y \sin(\theta)$
                \State $y' \gets -x \sin(\theta) + y \cos(\theta)$
                \State $kernel \gets \exp\left(-\frac{x'^2 + \gamma^2 y'^2}{2\sigma^2}\right) \cos\left(2\pi \frac{x'}{\lambda} + \psi\right)$
            \EndFor
        \EndFor
        \State $kernel \gets kernel - \text{mean}(kernel)$ 
        \State $scale \gets \text{Set fixed parameter}$
        \State $kernel \gets kernel \times scale$
        \State $weight[oc] \gets kernel$
    \EndFor
    \State \Return $weight$ as torch tensor
\EndWhile
\end{algorithmic}
\end{algorithm}

\subsection{Loss Function}
We use $L_1$ loss as the reconstruction loss:
\begin{equation}
    \mathcal{L}_{rec} = \| G - H \|_1,
\end{equation}
where $G$ is the ground truth image, and $H$ denotes the reconstructed image. 

\section{Experiments}
In this section, we evaluate our S2BNet on three pan-sharpening datasets. We also conduct extensive experiments to analyze our proposed model.

\subsection{Experimental Settings}

\textbf{Datasets.} For the pan-sharpening, three $4$-band datasets acquired by WorldView-2, GaoFen-2 and QuickBird sensors are adopted for experimental analysis. Due to the unavailability of ground-truth (GT) images for training, following Wald's protocol~\cite{wald1997fusion}. We employ downsampling operations to produce corresponding dataset for each satellite sensor, as shown in the Appendix, we present the detailed information of traning dataset and testing dataset in the experiment.

\textbf{Training Details.} Moreover, the proposed network is built with PyTorch and trained on four NVIDIA RTX $A5000$ GPUs. Training is conducted with a batch size of $16$, and the Adam optimizer is adopted, setting $\beta_1=0.9$ and $\beta_2=0.999$ for its momentum coefficients.The initial learning rate is set to $1.5\times 10^{-3}$ and is multiplied by $0.85$ every $100$ epochs. The whole training procedure runs for $1500$ epochs before stopping.

\textbf{Metrics.} Following previous studies on pan-sharpening, five image quality assessment metrics~\cite{Yang_Cao_Xiao_Zhou_Liu_chen_Meng_2023} are employed for evaluation on reduced-resolution images, including spectral angle mapper (SAM), dimensionless global error in synthesis (ERGAS), the structural similarity (SSIM), the peak signal-to-noise ratio (PSNR), and the Q-index. Since there are not GT images available for the full-resolution dataset, we use three non-reference metrics to evaluate the performance of the model: Spectral Distortion Index ($D_\lambda$), Spatial Distortion Index ($D_s$), and No-Reference Quality (QNR)~\cite{alparone2008multispectral}. 

\textbf{Efficiency Evaluation.} Following previous BNN work~\cite{cai2023binarized}, the FLOPs per second for BNN is determined by $\mathrm{Flops}^{\mathrm{b}}=\mathrm{Flops}^{\mathbf{f}}/64$. The overall FLOPs is $\mathrm{Flops}^{\mathrm{b}}+\mathrm{Flops}^{\mathrm{f}}$. The parameters of BNN are calculated as $\mathrm{Params}^{\mathrm{b}} =\mathrm{Params}^{\mathrm{f}}/32$. The total number of parameters is $\mathrm{Params}^{\mathrm{b}}+\mathrm{Params}^{\mathrm{f}}$. 

\subsection{Compare with State-of-the-arts}
\textbf{Comparison Methods.} We compare our method with various full precision pan-sharpening networks including DMFNet~\cite{Yuan2019DMFNet}, PANFormer~\cite{zhou2022panformer}, MutInf~\cite{zhou2022spatial}, FAMENet~\cite{DBLP:conf/aaai/HeYLX0Z24}, CANNet~\cite{Duan2024CANNet}, UGCC~\cite{Zeng2025Cross-Modal} and MSCSCFormer~\cite{Ye2024MSCSCformer}. We also compare our spatial-spectral binarized convolution with other binarization methods, including BNN~\cite{Hubara2016BNN}, LCRBNN~\cite{shang2022lipschitz}, BiSRNet~\cite{cai2023binarized}, BBCU~\cite{Xia2022BBCU}, FABNet~\cite{Jiang2023FABNet}, IRNet~\cite{Qin2020IRNet} and E2FIF~\cite{Song2023E2FIF}.

\begin{table}[]
\caption{Quantitative comparison of our S2BNet with our full-precision and binary methods on the GaoFen-2 dataset.}
\centering
\setlength{\extrarowheight}{0pt}
\addtolength{\extrarowheight}{\aboverulesep}
\addtolength{\extrarowheight}{\belowrulesep}
\setlength{\aboverulesep}{0pt}
\setlength{\belowrulesep}{0pt}
\resizebox{\linewidth}{!}{
\begin{tabular}{c|c|cc|ccccc|ccc} 
\toprule
\rowcolor[rgb]{0.922,0.922,0.922} {\cellcolor[rgb]{0.922,0.922,0.922}}                           & {\cellcolor[rgb]{0.922,0.922,0.922}}                         & {\cellcolor[rgb]{0.922,0.922,0.922}}                            & {\cellcolor[rgb]{0.922,0.922,0.922}}                          & \multicolumn{5}{c|}{Reduced-Resolution}                                                  & \multicolumn{3}{c}{Full-Resolution}                  \\
\rowcolor[rgb]{0.922,0.922,0.922} \multirow{-2}{*}{{\cellcolor[rgb]{0.922,0.922,0.922}}Category} & \multirow{-2}{*}{{\cellcolor[rgb]{0.922,0.922,0.922}}Method} & \multirow{-2}{*}{{\cellcolor[rgb]{0.922,0.922,0.922}}Params (K)} & \multirow{-2}{*}{{\cellcolor[rgb]{0.922,0.922,0.922}}Flops (G)} & PSNR$\uparrow$             & SSIM$\uparrow$            & $Q_4\uparrow$               & SAM$\downarrow$             & ERGAS$\downarrow$           & $D_\lambda\downarrow$        & $D_s\downarrow$            & QNR$\uparrow$              \\ 
\hline
\rowcolor[rgb]{0.941,1,0.941} {\cellcolor[rgb]{0.941,1,0.941}}                                   & DMFNet ~\cite{Yuan2019DMFNet}                                                     & 1631.924                                                        & 145.811                                                       & 43.1223          & 0.9734          & 0.8557          & 0.0322          & 1.5660          & 0.0678          & 0.1134          & 0.8265           \\
\rowcolor[rgb]{0.941,1,0.941} {\cellcolor[rgb]{0.941,1,0.941}}                                   & PANFormer~\cite{zhou2022panformer}                                                   & 1530.300                                                          & 12.002                                                        & 44.8501          & 0.9805          & 0.8865          & 0.0271          & 1.3337          & 0.0670          & 0.1806          & 0.7639           \\
\rowcolor[rgb]{0.941,1,0.941} {\cellcolor[rgb]{0.941,1,0.941}}                                   & MutInf~\cite{zhou2022spatial}                                                       & 185.496                                                         & 9.986                                                         & 44.8306          & 0.9800          & 0.8836          & 0.0277          & 1.3394          & 0.0755          & 0.1762          & 0.7613           \\
\rowcolor[rgb]{0.941,1,0.941} {\cellcolor[rgb]{0.941,1,0.941}}                                   & FAMENet~\cite{DBLP:conf/aaai/HeYLX0Z24}                                                      & 1244.228                                                        & 39.272                                                        & 45.6617          & 0.9837          & 0.8966          & 0.0248          & 1.2142          & 0.0697          & 0.1800          & 0.7622           \\
\rowcolor[rgb]{0.941,1,0.941} {\cellcolor[rgb]{0.941,1,0.941}}                                   & CANNet~\cite{Duan2024CANNet}                                                        & 785.118                                                         & 3.237                                                         & 45.2222          & 0.9816          & 0.8869          & 0.0267          & 1.2748          & 0.0654          & 0.1969          & 0.7498           \\
\rowcolor[rgb]{0.941,1,0.941} {\cellcolor[rgb]{0.941,1,0.941}}                                   & UGCC ~\cite{Zeng2025Cross-Modal}                                                        & 233.233                                                         & 307.495                                                       & 40.5620          & 0.9586          & 0.7803          & 0.0448          & 2.1717          & 0.0893          & 0.0608          & 0.8564           \\
\rowcolor[rgb]{0.941,1,0.941} \multirow{-7}{*}{{\cellcolor[rgb]{0.941,1,0.941}}Full}            & MSCSCFormer~\cite{Ye2024MSCSCformer}                                                   & 1950.686                                                        & 582.435                                                       & 42.6871          & 0.9696          & 0.8350          & 0.0360          & 1.7149          & 0.0692          & 0.0973          & 0.8403           \\ 
\hline
\rowcolor[rgb]{1,0.969,0.969} {\cellcolor[rgb]{1,0.961,0.961}}                                   & BNN ~\cite{Hubara2016BNN}                                                          & 17.760                                                          & 1.938                                                         & 34.4931          & 0.8802          & 0.4701          & 0.0860          & 4.1576          & 0.0985          & 0.0871          & 0.8243           \\
\rowcolor[rgb]{1,0.969,0.969} {\cellcolor[rgb]{1,0.961,0.961}}                                   & LCRBNN ~\cite{shang2022lipschitz}                                                       & 17.760                                                          & 1.938                                                         & 32.2058          & 0.8165          & 0.2852          & 0.0952          & 5.4921          & 0.1043          & 0.1621          & 0.7503           \\
\rowcolor[rgb]{1,0.969,0.969} {\cellcolor[rgb]{1,0.961,0.961}}                                   & BiSRNet ~\cite{cai2023binarized}                                                      & 26.459                                                          & 0.664                                                         & 43.3446          & 0.9727          & 0.8424          & 0.0327          & 1.5748          & 0.0669          & 0.0781          & 0.8604           \\
\rowcolor[rgb]{1,0.969,0.969} {\cellcolor[rgb]{1,0.961,0.961}}                                   & BBCU ~\cite{Xia2022BBCU}                                                         & 23.520                                                          & 0.664                                                         & 43.0233          & 0.9712          & 0.8361          & 0.0335          & 1.6449          & 0.0676          & 0.0900          & 0.8484           \\
\rowcolor[rgb]{1,0.969,0.969} {\cellcolor[rgb]{1,0.961,0.961}}                                   & FABNet ~\cite{Jiang2023FABNet}                                                       & 59.296                                                          & 0.664                                                         & 43.0144          & 0.9709          & 0.8370          & 0.0342          & 1.6477          & 0.0676          & 0.0867          & 0.8515           \\
\rowcolor[rgb]{1,0.969,0.969} {\cellcolor[rgb]{1,0.961,0.961}}                                   & IRNet ~\cite{Qin2020IRNet}                                                        & 17.760                                                          & 1.938                                                         & 42.8422          & 0.9714          & 0.8394          & 0.0338          & 1.6636          & 0.0692          & 0.1095          & 0.8285           \\
\rowcolor[rgb]{1,0.969,0.969} {\cellcolor[rgb]{1,0.961,0.961}}                                   & E2FIF ~\cite{Song2023E2FIF}                                                        & 20.640                                                          & 0.664                                                         & 41.0767          & 0.9544          & 0.7992          & 0.0402          & 2.0694          & 0.0740          & 0.0909          & 0.8422           \\
\rowcolor[rgb]{1,0.969,0.969} \multirow{-8}{*}{{\cellcolor[rgb]{1,0.961,0.961}}Binary}           & \textbf{S2BNet}                                                         & 75.407                                                          & 0.920                                                         & \textbf{43.5619} & \textbf{0.9745} & \textbf{0.8523} & \textbf{0.0317} & \textbf{1.5384} & \textbf{0.0655} & \textbf{0.0680} & \textbf{0.8714}  \\
\bottomrule
\end{tabular}
}
\label{tab:gf2}
\end{table}

\textbf{Quantitative Comparison.} As summarized in Tab.~\ref{tab:gf2}, our method delivers substantial performance improvements over other binary techniques across all metrics, showcasing superior spectral and spatial fidelity. Lower SAM values demonstrate its effective spectral preservation, while higher PSNR scores confirm its robust spatial detail retention. More importantly, our model surpasses the majority of full-precision baselines, demonstrating its promising expressiveness while benefiting from the efficiency of model binarization.

We additionally assess our model's generalization capability in real-world scenarios by testing it on full-resolution data. As shown in Tab.~\ref{tab:gf2}, our model consistently achieves superior performance across all three datasets, yielding the optimal scores on most evaluation metrics. The quantitative results on the WorldView-2 and QuickBird examples are shown in Appendix.

\begin{figure}[]
    \centering
    \includegraphics[width=\linewidth]{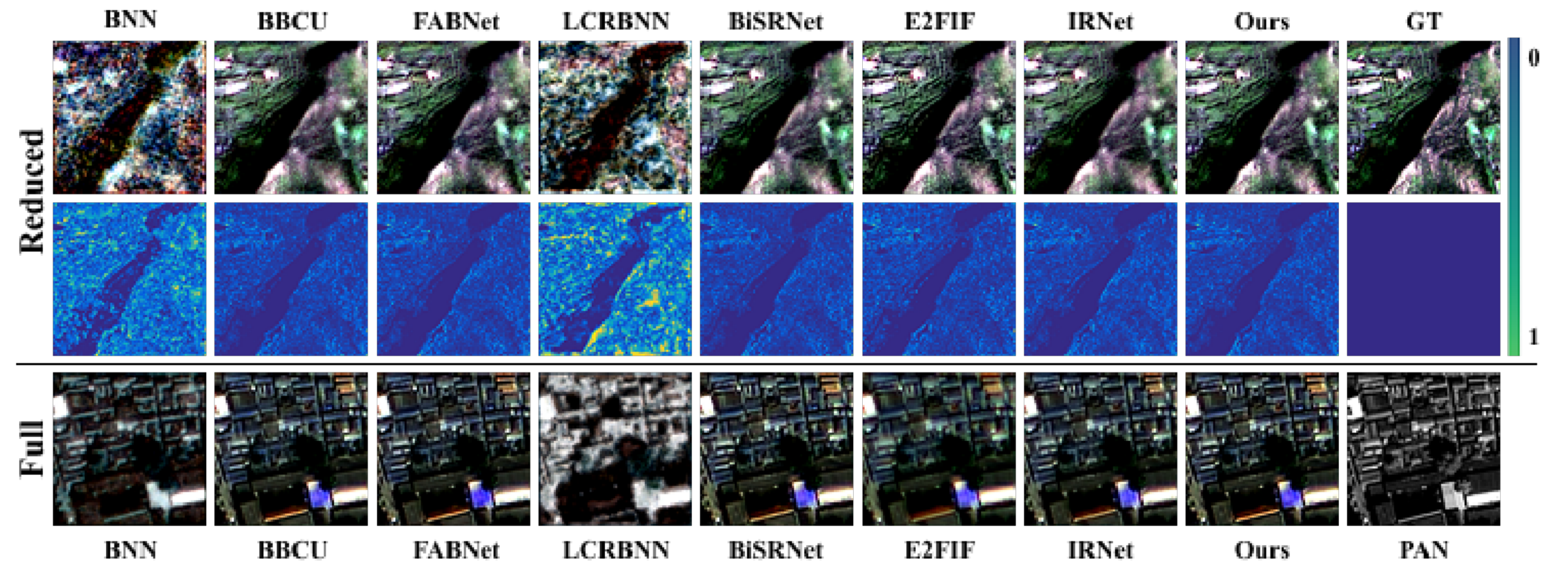}
    \caption{Visual comparison between our model and other binary methods on GF-2 example. The top two lines represent the reconstructed results and corresponding MAE maps of the reduced-resolution example, and the last line represents the reconstructed results of the full-resolution example.}
    \label{fig:gf2}
\end{figure}

\textbf{Visual Comparison.} The qualitative results on the GaoFen-2 dataset are illustrated in Fig.~\ref{fig:gf2}. It can be observed that our model produces image with clear textures and visually pleasing spectra, and the corresponding MAE map exhibits minimal bright spots, indicating a high resemblance to the ground truth. In addition, the image generated by our S2BNet exhibits reduced aberrations and artifacts in full-resolution example. The visualization results on the WorldView-2 example are shown in Appendix.

\textbf{Performance-Efficiency Comparison.} The performance-efficiency comparison between our method and other full-precision and binary benchmarks is also presented in Tab.~\ref{tab:gf2}. Compared with full-precision methods, our approach not only surpasses most methods in performance but also minimizes computational overhead. Compared to other binary models, our S2BNet achieves the highest PSNR score, with a marginally lower FLOPs than IRNet.

\subsection{Ablation Study}

\textbf{Effect of the Spectral Redistribution Mechanism.} To verify the effect of Spectral-Redistribution Mechanism (SRM) in the S2B-Conv, Config (II) in Tab.~\ref{tab:config} is obtained by removing Spectral Redistribution Mechanism, and uses random initial scaling and bias values. It could be reported in Tab.~\ref{tab:config} that the model achieves $0.31$ dB improvement when we exploit SRM.

\textbf{Effect of the Gabor Spatial Feature Amplifier.} As shown in Tab.~\ref{tab:config}, the Config (I) is denoted, which replaces the Gabor Spatial Feature Amplifier (GSFA) with traditional convolutional weights. Our GSFA dramatically surpasses the common convolutional weights by $0.46$ dB.

\textbf{Binarizing Different Parts.} We binarize one part of the S2BNet while keeping the other parts full-precision to study the binarization benefit. The results are reported in Tab.~\ref{tab:binarize}. The base model (full precision) yields 42.2759 dB in PSNR while costing 13.884 G OPs and 349.695 K Params. We can find that (i) Binarizing the bottleneck reduces the Params the most (100.329K) with the smallest performance drop (only 0.16 dB). (ii) Binarizing the decoder achieves the largest OPs reduction (6.177G) while the performance degrades by a moderate margin (0.57 dB).

\begin{table}[ht]
\centering
\begin{minipage}[t]{0.38\textwidth}
\centering
\caption{Comparison of different configurations.}
\label{tab:config}
\centering
\setlength{\extrarowheight}{0pt}
\addtolength{\extrarowheight}{\aboverulesep}
\addtolength{\extrarowheight}{\belowrulesep}
\setlength{\aboverulesep}{0pt}
\setlength{\belowrulesep}{0pt}
\resizebox{\linewidth}{!}{
\begin{tabular}{ccccccc} 
\toprule
Config  &  GSFA & SRM           & PSNR$\uparrow$    & SSIM$\uparrow$   & SAM$\downarrow$      & QNR$\uparrow$     \\ 
\midrule
(I) &  \XSolid    & \Checkmark                  & 43.107 & 0.972 & 0.033   & 0.851  \\
(II)  &   \Checkmark       &    \XSolid                & 43.250 & 0.972 & 0.033   & 0.859  \\
\rowcolor[rgb]{0.922,0.922,0.922}
Ours &  \Checkmark       & \Checkmark                  & \textbf{43.562} & \textbf{0.975} & \textbf{0.031} & \textbf{0.871}  \\
\bottomrule
\end{tabular}
}
\end{minipage}
\hfill
\begin{minipage}[t]{0.60\textwidth}
\caption{Ablation study of binarizing different parts of the base model.}
\label{tab:binarize}
\centering
\setlength{\extrarowheight}{0pt}
\addtolength{\extrarowheight}{\aboverulesep}
\addtolength{\extrarowheight}{\belowrulesep}
\setlength{\aboverulesep}{0pt}
\setlength{\belowrulesep}{0pt}
\resizebox{\linewidth}{!}{
\begin{tabular}{cccccccc} 
\toprule
           & $\mathbf{Flops}^{\mathbf{f}}$ & $\mathbf{Flops}^{\mathbf{b}}$ & $\mathbf{Params}^{\mathbf{f}}$ & $\mathbf{Params}^{\mathbf{b}}$  & $\mathbf{Flops}^{\mathbf{t}}$   & $\mathbf{Params}^{\mathbf{t}}$  & PSNR$\uparrow$  \\ 
\midrule
Encoder    & 4.194    & 0.066    & 85.504     & 2.672      & 9.755  & 266.863 &   41.3088    \\
Bottelneck & 1.493    & 0.023    & 100.352    & 3.136      & 12.414 & 252.479 &   42.1118    \\
Decoder    & 6.275    & 0.098    & 88.064     & 2.752      & 7.707  & 264.383 &    41.7051   \\
\bottomrule
\end{tabular}
}
\end{minipage}
\end{table}

\section{Conclusion}
In this paper, we propose a novel BNN-based method, S2BNet, for pan-sharpening. S2BNet is the compact and easy-to-deploy base model with simple computation operations. Specifically, we customize the basic unit S2B-Conv for model binarization. S2B-Conv utilizes the Spectral-Redistribution Mechanism, which can adaptively adjust the density and distribution of spectral features. Besides, S2B-Conv employs the Gabor Spatial Feature Amplifier to capture multi-scale and multi-directional spatial features. Comprehensive quantitative and qualitative experiments demonstrate that our S2BNet significantly outperforms SOTA BNNs and even achieves comparable performance with full-precision pan-sharpening algorithms.


\section{Reproducibility Statement}
To ensure the reproducibility of our research, we will make the core code publicly available in the supplementary materials.

\bibliography{iclr2026_conference}

\begin{thebibliography}{45}
\providecommand{\natexlab}[1]{#1}
\providecommand{\url}[1]{\texttt{#1}}
\expandafter\ifx\csname urlstyle\endcsname\relax
  \providecommand{\doi}[1]{doi: #1}\else
  \providecommand{\doi}{doi: \begingroup \urlstyle{rm}\Url}\fi

\bibitem[Alparone et~al.(2008)Alparone, Aiazzi, Baronti, Garzelli, Nencini, and Selva]{alparone2008multispectral}
Luciano Alparone, Bruno Aiazzi, Stefano Baronti, Andrea Garzelli, Filippo Nencini, and Massimo Selva.
\newblock Multispectral and panchromatic data fusion assessment without reference.
\newblock \emph{Photogrammetric Engineering \& Remote Sensing}, 74\penalty0 (2):\penalty0 193--200, 2008.

\bibitem[Asokan \& Anitha(2019)Asokan and Anitha]{Asokan_Anitha_2019}
Anju Asokan and J.~Anitha.
\newblock Change detection techniques for remote sensing applications: a survey.
\newblock \emph{Earth Science Informatics}, pp.\  143–160, Jun 2019.

\bibitem[Bandara \& Patel(2022{\natexlab{a}})Bandara and Patel]{DBLP:conf/cvpr/BandaraP22}
Wele Gedara~Chaminda Bandara and Vishal~M. Patel.
\newblock Hypertransformer: {A} textural and spectral feature fusion transformer for pansharpening.
\newblock In \emph{{IEEE/CVF} Conference on Computer Vision and Pattern Recognition, {CVPR} 2022, New Orleans, LA, USA, June 18-24, 2022}, pp.\  1757--1767. {IEEE}, 2022{\natexlab{a}}.

\bibitem[Bandara \& Patel(2022{\natexlab{b}})Bandara and Patel]{bandara2022hypertransformer}
Wele Gedara~Chaminda Bandara and Vishal~M Patel.
\newblock Hypertransformer: A textural and spectral feature fusion transformer for pansharpening.
\newblock In \emph{Proceedings of the IEEE/CVF conference on computer vision and pattern recognition}, pp.\  1767--1777, 2022{\natexlab{b}}.

\bibitem[Bengio et~al.(2013)Bengio, L{\'e}onard, and Courville]{bengio2013estimating}
Yoshua Bengio, Nicholas L{\'e}onard, and Aaron Courville.
\newblock Estimating or propagating gradients through stochastic neurons for conditional computation.
\newblock \emph{arXiv preprint arXiv:1308.3432}, 2013.

\bibitem[Cai \& Huang(2020)Cai and Huang]{cai2020super}
Jiajun Cai and Bo~Huang.
\newblock Super-resolution-guided progressive pansharpening based on a deep convolutional neural network.
\newblock \emph{IEEE Transactions on Geoscience and Remote Sensing}, 59\penalty0 (6):\penalty0 5206--5220, 2020.

\bibitem[Cai et~al.(2023)Cai, Zheng, Lin, Wang, Yuan, and Zhang]{cai2023binarized}
Yuanhao Cai, Yuxin Zheng, Jing Lin, Haoqian Wang, Xin Yuan, and Yulun Zhang.
\newblock Binarized spectral compressive imaging, 2023.

\bibitem[Duan et~al.(2024)Duan, Wu, Deng, and Deng]{Duan2024CANNet}
Yule Duan, Xiao Wu, Haoyu Deng, and Liang-Jian Deng.
\newblock Content-adaptive non-local convolution for remote sensing pansharpening.
\newblock In \emph{2024 IEEE/CVF Conference on Computer Vision and Pattern Recognition (CVPR)}, pp.\  27738--27747, 2024.

\bibitem[Fasbender et~al.(2008)Fasbender, Radoux, and Bogaert]{Fasbender_Radoux_Bogaert_2008}
D.~Fasbender, J.~Radoux, and P.~Bogaert.
\newblock Bayesian data fusion for adaptable image pansharpening.
\newblock \emph{IEEE Transactions on Geoscience and Remote Sensing}, pp.\  1847–1857, Jun 2008.

\bibitem[Gillespie et~al.(1987)Gillespie, Kahle, and Walker]{Gillespie_Kahle_Walker_1987}
Alan~R Gillespie, Anne~B Kahle, and Richard~E Walker.
\newblock Color enhancement of highly correlated images. ii. channel ratio and “chromaticity” transformation techniques.
\newblock \emph{Remote Sensing of Environment}, pp.\  343–365, Jul 1987.

\bibitem[Han et~al.(2024)Han, Xu, Gao, Li, and Zhang]{han2024gretnet}
Zhu Han, Shuyi Xu, Lianru Gao, Zhi Li, and Bing Zhang.
\newblock Gretnet: Gaussian retentive network for hyperspectral image classification.
\newblock \emph{IEEE Geoscience and Remote Sensing Letters}, 2024.

\bibitem[Hou et~al.(2025)Hou, Chen, Ran, Cong, Liu, You, and Deng]{hou2025binarized}
Junming Hou, Xiaoyu Chen, Ran Ran, Xiaofeng Cong, Xinyang Liu, Jian~Wei You, and Liang-Jian Deng.
\newblock Binarized neural network for multi-spectral image fusion.
\newblock In \emph{Proceedings of the Computer Vision and Pattern Recognition Conference}, pp.\  2236--2245, 2025.

\bibitem[Hubara et~al.(2016{\natexlab{a}})Hubara, Courbariaux, Soudry, El-Yaniv, and Bengio]{Hubara2016BNN}
Itay Hubara, Matthieu Courbariaux, Daniel Soudry, Ran El-Yaniv, and Yoshua Bengio.
\newblock Binarized neural networks.
\newblock In \emph{Proceedings of the 30th International Conference on Neural Information Processing Systems}, NIPS'16, pp.\  4114–4122, Red Hook, NY, USA, 2016{\natexlab{a}}. Curran Associates Inc.

\bibitem[Hubara et~al.(2016{\natexlab{b}})Hubara, Courbariaux, Soudry, El-Yaniv, and Bengio]{hubara2016binarized}
Itay Hubara, Matthieu Courbariaux, Daniel Soudry, Ran El-Yaniv, and Yoshua Bengio.
\newblock Binarized neural networks.
\newblock \emph{Advances in neural information processing systems}, 29, 2016{\natexlab{b}}.

\bibitem[Jiang et~al.(2023)Jiang, Wang, Xin, Li, Yang, Li, Wang, and Gao]{Jiang2023FABNet}
Xinrui Jiang, Nannan Wang, Jingwei Xin, Keyu Li, Xi~Yang, Jie Li, Xiaoyu Wang, and Xinbo Gao.
\newblock Fabnet: Frequency-aware binarized network for single image super-resolution.
\newblock \emph{IEEE Transactions on Image Processing}, 32:\penalty0 6234--6247, 2023.

\bibitem[Li et~al.(2022)Li, Zheng, Yao, Gao, and Hong]{li2022deep}
Jiaxin Li, Ke~Zheng, Jing Yao, Lianru Gao, and Danfeng Hong.
\newblock Deep unsupervised blind hyperspectral and multispectral data fusion.
\newblock \emph{IEEE Geoscience and Remote Sensing Letters}, 19:\penalty0 1--5, 2022.

\bibitem[Li et~al.(2023{\natexlab{a}})Li, Liu, Xiao, Huang, and Yang]{DBLP:conf/ijcai/LiLXHY23}
Mingsong Li, Yikun Liu, Tao Xiao, Yuwen Huang, and Gongping Yang.
\newblock Local-global transformer enhanced unfolding network for pan-sharpening.
\newblock In \emph{Proceedings of the Thirty-Second International Joint Conference on Artificial Intelligence, {IJCAI} 2023, 19th-25th August 2023, Macao, SAR, China}, pp.\  1071--1079. ijcai.org, 2023{\natexlab{a}}.

\bibitem[Li et~al.(2023{\natexlab{b}})Li, Liu, Xiao, Huang, and Yang]{li2023local}
Mingsong Li, Yikun Liu, Tao Xiao, Yuwen Huang, and Gongping Yang.
\newblock Local-global transformer enhanced unfolding network for pan-sharpening.
\newblock \emph{arXiv preprint arXiv:2304.14612}, 2023{\natexlab{b}}.

\bibitem[Liu et~al.(2018)Liu, Wu, Luo, Yang, Liu, and Cheng]{liu2018bi}
Zechun Liu, Baoyuan Wu, Wenhan Luo, Xin Yang, Wei Liu, and Kwang-Ting Cheng.
\newblock Bi-real net: Enhancing the performance of 1-bit cnns with improved representational capability and advanced training algorithm.
\newblock In \emph{Proceedings of the European conference on computer vision (ECCV)}, pp.\  722--737, 2018.

\bibitem[Ma et~al.(2025)Ma, Jiang, Zhao, Ma, Zhang, and Song]{ma2025deep}
Mengting Ma, Yizhen Jiang, Mengjiao Zhao, Xiaowen Ma, Wei Zhang, and Siyang Song.
\newblock Deep spatial--spectral fusion transformer for remote sensing pansharpening.
\newblock \emph{Information Fusion}, 118:\penalty0 102980, 2025.

\bibitem[Masi et~al.(2016)Masi, Cozzolino, Verdoliva, and Scarpa]{Masi_Cozzolino_Verdoliva_Scarpa_2016}
Giuseppe Masi, Davide Cozzolino, Luisa Verdoliva, and Giuseppe Scarpa.
\newblock Pansharpening by convolutional neural networks.
\newblock \emph{Remote Sensing}, 2016.
\newblock \doi{10.3390/rs8070594}.
\newblock URL \url{http://dx.doi.org/10.3390/rs8070594}.

\bibitem[Nunez et~al.(1999)Nunez, Otazu, Fors, Prades, Pala, and Arbiol]{Nunez_Otazu_Fors_Prades_Pala_Arbiol_1999}
J.~Nunez, X.~Otazu, O.~Fors, A.~Prades, V.~Pala, and R.~Arbiol.
\newblock Multiresolution-based image fusion with additive wavelet decomposition.
\newblock \emph{IEEE Transactions on Geoscience and Remote Sensing}, pp.\  1204–1211, May 1999.

\bibitem[Qin et~al.(2020{\natexlab{a}})Qin, Gong, Liu, Bai, Song, and Sebe]{qin2020binary}
Haotong Qin, Ruihao Gong, Xianglong Liu, Xiao Bai, Jingkuan Song, and Nicu Sebe.
\newblock Binary neural networks: A survey.
\newblock \emph{Pattern Recognition}, 105:\penalty0 107281, 2020{\natexlab{a}}.

\bibitem[Qin et~al.(2020{\natexlab{b}})Qin, Gong, Liu, Shen, Wei, Yu, and Song]{Qin2020IRNet}
Haotong Qin, Ruihao Gong, Xianglong Liu, Mingzhu Shen, Ziran Wei, Fengwei Yu, and Jingkuan Song.
\newblock Forward and backward information retention for accurate binary neural networks.
\newblock In \emph{2020 IEEE/CVF Conference on Computer Vision and Pattern Recognition (CVPR)}, pp.\  2247--2256, 2020{\natexlab{b}}.

\bibitem[Rastegari et~al.(2016)Rastegari, Ordonez, Redmon, and Farhadi]{rastegari2016xnor}
Mohammad Rastegari, Vicente Ordonez, Joseph Redmon, and Ali Farhadi.
\newblock Xnor-net: Imagenet classification using binary convolutional neural networks.
\newblock In \emph{European conference on computer vision}, pp.\  525--542. Springer, 2016.

\bibitem[Shang et~al.(2022)Shang, Xu, Duan, Zong, Nie, and Yan]{shang2022lipschitz}
Yuzhang Shang, Dan Xu, Bin Duan, Ziliang Zong, Liqiang Nie, and Yan Yan.
\newblock Lipschitz continuity retained binary neural network.
\newblock In \emph{European conference on computer vision}, pp.\  603--619. Springer, 2022.

\bibitem[Song et~al.(2023)Song, Lang, Wei, and Zhang]{Song2023E2FIF}
Chongxing Song, Zhiqiang Lang, Wei Wei, and Lei Zhang.
\newblock E2fif: Push the limit of binarized deep imagery super-resolution using end-to-end full-precision information flow.
\newblock \emph{IEEE Transactions on Image Processing}, 32:\penalty0 5379--5393, 2023.

\bibitem[Tan et~al.(2024)Tan, Huang, Zheng, Zhou, Yan, Hong, and Zhao]{tan2024revisiting}
Jiangtong Tan, Jie Huang, Naishan Zheng, Man Zhou, Keyu Yan, Danfeng Hong, and Feng Zhao.
\newblock Revisiting spatial-frequency information integration from a hierarchical perspective for panchromatic and multi-spectral image fusion.
\newblock In \emph{Proceedings of the IEEE/CVF Conference on Computer Vision and Pattern Recognition}, pp.\  25922--25931, 2024.

\bibitem[Wald et~al.(1997)Wald, Ranchin, and Mangolini]{wald1997fusion}
Lucien Wald, Thierry Ranchin, and Marc Mangolini.
\newblock Fusion of satellite images of different spatial resolutions: Assessing the quality of resulting images.
\newblock \emph{Photogrammetric engineering and remote sensing}, 63\penalty0 (6):\penalty0 691--699, 1997.

\bibitem[Wang \& Fu(2022)Wang and Fu]{wang2022trainability}
Huan Wang and Yun Fu.
\newblock Trainability preserving neural pruning.
\newblock \emph{arXiv preprint arXiv:2207.12534}, 2022.

\bibitem[Wang et~al.(2022)Wang, Lohit, Jones, and Fu]{wang2022makes}
Huan Wang, Suhas Lohit, Michael~N Jones, and Yun Fu.
\newblock What makes a" good" data augmentation in knowledge distillation-a statistical perspective.
\newblock \emph{Advances in Neural Information Processing Systems}, 35:\penalty0 13456--13469, 2022.

\bibitem[Wu et~al.(2022)Wu, Huang, Deng, and Zhang]{Wu_Huang_Deng_Zhang_2022}
Xiao Wu, Ting-Zhu Huang, Liang-Jian Deng, and Tian-Jing Zhang.
\newblock Dynamic cross feature fusion for remote sensing pansharpening.
\newblock In \emph{2021 IEEE/CVF International Conference on Computer Vision (ICCV)}, Mar 2022.

\bibitem[Xia et~al.(2022)Xia, Zhang, Wang, Tian, Wenming, Timofte, and Gool]{Xia2022BBCU}
Bin Xia, Yulun Zhang, Yitong Wang, Yapeng Tian, Yang Wenming, Radu Timofte, and Luc Gool.
\newblock Basic binary convolution unit for binarized image restoration network.
\newblock 10 2022.
\newblock \doi{10.48550/arXiv.2210.00405}.

\bibitem[Xin et~al.(2020)Xin, Wang, Jiang, Li, Huang, and Gao]{xin2020binarized}
Jingwei Xin, Nannan Wang, Xinrui Jiang, Jie Li, Heng Huang, and Xinbo Gao.
\newblock Binarized neural network for single image super resolution.
\newblock In \emph{European conference on computer vision}, pp.\  91--107. Springer, 2020.

\bibitem[Xing et~al.(2023)Xing, Zhang, He, Zhang, and Zhang]{Xing_Zhang_He_Zhang_Zhang_2023}
Yinghui Xing, Yan Zhang, Houjun He, Xiuwei Zhang, and Yanning Zhang.
\newblock Pansharpening via frequency-aware fusion network with explicit similarity constraints.
\newblock \emph{IEEE Transactions on Geoscience and Remote Sensing}, pp.\  1–1, Jan 2023.

\bibitem[Xuanhua et~al.(2024)Xuanhua, Keyu, Rui, Chengjun, Jie, and Man]{DBLP:conf/aaai/HeYLX0Z24}
He~Xuanhua, Yan Keyu, Li~Rui, Xie Chengjun, Zhang Jie, and Zhou Man.
\newblock Frequency-adaptive pan-sharpening with mixture of experts.
\newblock In \emph{2024 Conference on Innovative Applications of Artificial Intelligence (AAAI)}, 2024.

\bibitem[Yang et~al.(2023)Yang, Cao, Xiao, Zhou, Liu, chen, and Meng]{Yang_Cao_Xiao_Zhou_Liu_chen_Meng_2023}
Gang Yang, Xiangyong Cao, Wenzhe Xiao, Man Zhou, Aiping Liu, Xun chen, and Deyu Meng.
\newblock Panflownet: A flow-based deep network for pan-sharpening.
\newblock May 2023.

\bibitem[Yang et~al.(2017)Yang, Fu, Hu, Huang, Ding, and Paisley]{Yang_Fu_Hu_Huang_Ding_Paisley_2017}
Junfeng Yang, Xueyang Fu, Yuwen Hu, Yue Huang, Xinghao Ding, and John Paisley.
\newblock Pannet: A deep network architecture for pan-sharpening.
\newblock In \emph{2017 IEEE International Conference on Computer Vision (ICCV)}, Oct 2017.
\newblock \doi{10.1109/iccv.2017.193}.
\newblock URL \url{http://dx.doi.org/10.1109/iccv.2017.193}.

\bibitem[Ye et~al.(2024)Ye, Wang, Fang, and Zhang]{Ye2024MSCSCformer}
Yongxu Ye, Tingting Wang, Faming Fang, and Guixu Zhang.
\newblock Mscscformer: Multiscale convolutional sparse coding-based transformer for pansharpening.
\newblock \emph{IEEE Transactions on Geoscience and Remote Sensing}, 62:\penalty0 1--12, 2024.

\bibitem[Yuan et~al.(2019)Yuan, Zhou, and Luo]{Yuan2019DMFNet}
Jianzhong Yuan, Wujie Zhou, and Ting Luo.
\newblock Dmfnet: Deep multi-modal fusion network for rgb-d indoor scene segmentation.
\newblock \emph{IEEE Access}, 7:\penalty0 169350--169358, 2019.
\newblock \doi{10.1109/ACCESS.2019.2955101}.

\bibitem[Zeng et~al.(2025)Zeng, Yang, Shen, Li, Jiang, and Li]{Zeng2025Cross-Modal}
Haoying Zeng, Xiaoyuan Yang, Kangqing Shen, Yixiao Li, Jin Jiang, and Fangyi Li.
\newblock Cross-modal contrastive pansharpening via uncertainty guidance.
\newblock \emph{IEEE Transactions on Geoscience and Remote Sensing}, 63:\penalty0 1--14, 2025.

\bibitem[Zhang et~al.(2024)Zhang, Zhang, Yuan, and Fu]{zhang2024binarized}
Gengchen Zhang, Yulun Zhang, Xin Yuan, and Ying Fu.
\newblock Binarized low-light raw video enhancement.
\newblock In \emph{Proceedings of the IEEE/CVF Conference on Computer Vision and Pattern Recognition}, pp.\  25753--25762, 2024.

\bibitem[Zhou et~al.(2022{\natexlab{a}})Zhou, Liu, and Wang]{zhou2022panformer}
Huanyu Zhou, Qingjie Liu, and Yunhong Wang.
\newblock Panformer: A transformer based model for pan-sharpening.
\newblock In \emph{2022 IEEE International Conference on Multimedia and Expo (ICME)}, pp.\  1--6. IEEE, 2022{\natexlab{a}}.

\bibitem[Zhou et~al.(2022{\natexlab{b}})Zhou, Huang, Fang, Fu, and Liu]{zhou2022pan}
Man Zhou, Jie Huang, Yanchi Fang, Xueyang Fu, and Aiping Liu.
\newblock Pan-sharpening with customized transformer and invertible neural network.
\newblock In \emph{Proceedings of the AAAI conference on artificial intelligence}, volume~36, pp.\  3553--3561, 2022{\natexlab{b}}.

\bibitem[Zhou et~al.(2022{\natexlab{c}})Zhou, Huang, Yan, Yu, Fu, Liu, Wei, and Zhao]{zhou2022spatial}
Man Zhou, Jie Huang, Keyu Yan, Hu~Yu, Xueyang Fu, Aiping Liu, Xian Wei, and Feng Zhao.
\newblock Spatial-frequency domain information integration for pan-sharpening.
\newblock In \emph{European conference on computer vision}, pp.\  274--291. Springer, 2022{\natexlab{c}}.

\end{thebibliography}
\bibliographystyle{iclr2026_conference}

\appendix
\section{The use of LLMS}
We utilized GPT and Kimi to correct grammatical errors and translate some sentences.

\section{Appendix}
In this Appendix, we provide more experimental details of our S2BNet as follows.

\textbf{Pan-sharpening Datasets.} We conduct experiments using the widely recognized WorldView-2, QuickBird and GaoFen-2 datasets~\cite{li2023local}. The WorldView-2 dataset consists of instances acquired by the sensor aboard the WorldView-2 satellite. This sensor captures data which covers wavelengths from $0.4$ to $1$ $\mathrm{\mu m}$, with a spatial resolution of $1.24$ $\mathrm{m}$. The QuickBird dataset consists of instances acquired by the sensor aboard the QuickBird satellite. This sensor captures data across four spectral bands, covering wavelengths from $0.45$ to $0.9$ $\mathrm{\mu m}$, with a spatial resolution of $2.4$ $\mathrm{m}$. The images in the GaoFen-2 dataset are collected by the sensor onboard the GaoFen-2 satellite, which records data across four spectral bands within the wavelength range of $0.45-0.89$ $\mathrm{\mu m}$. Additionally, this sensor provides a spatial resolution of $3.2m$. As shown in Tab.~\ref{tab:data}, we present the detailed information of training dataset and the testing dataset in the experiment.

\begin{table*}[!h]
\caption{The detailed dataset information (GaoFen-2, WorldView-2 and QuickBird), followed by~\cite{ma2025deep}.}
\centering
    \setlength{\tabcolsep}{10pt}
    \resizebox{1\linewidth}{!}{ 
    \begin{tabular}{cccccc}
        \specialrule{1pt}{0pt}{0pt}
        \textbf{Sensor} & \textbf{ bit depth} & \textbf{\#Images} & \textbf{Scale} & \textbf{Training} & \textbf{Testing} \\
        \hline
         \multicolumn{1}{c}{\multirow{8}{*}{WordView-2}} & \multicolumn{1}{c}{\multirow{8}{*}{11}} & \multicolumn{1}{c}{\multirow{8}{*}{1}} & \multicolumn{1}{c}{\multirow{4}{*}{Reduced}} & \textbf{\#Patches:} 1012 & \textbf{\#Patches:} 145 \\
         \multicolumn{1}{c}{} & \multicolumn{1}{c}{} & \multicolumn{1}{c}{} & \multicolumn{1}{c}{} & \textbf{PAN:} 128 $ \times $ 128 $ \times $ 1 & \textbf{PAN:} 128 $ \times $ 128 $ \times $ 1 \\
         \multicolumn{1}{c}{} & \multicolumn{1}{c}{} & \multicolumn{1}{c}{} & \multicolumn{1}{c}{} & \textbf{LR-MS:} 32 $ \times $ 32 $ \times $ 4 & \textbf{LR-MS:} 32 $ \times $ 32 $ \times $ 4 \\
         \multicolumn{1}{c}{} & \multicolumn{1}{c}{} & \multicolumn{1}{c}{} & \multicolumn{1}{c}{} & \textbf{Output:} 128 $ \times $ 128 $ \times $ 4 & \textbf{Output:} 128 $ \times $ 128 $ \times $ 4 \\
         \cline{4-6}
         \multicolumn{1}{c}{} & \multicolumn{1}{c}{}  & \multicolumn{1}{c}{} & \multicolumn{1}{c}{\multirow{4}{*}{Full}} & - & \textbf{\#Patches:} 120 \\
         \multicolumn{1}{c}{} & \multicolumn{1}{c}{}  & \multicolumn{1}{c}{} & \multicolumn{1}{c}{} &-& \textbf{PAN:} 128 $ \times $ 128 $ \times $ 1 \\
         \multicolumn{1}{c}{} & \multicolumn{1}{c}{}  & \multicolumn{1}{c}{} & \multicolumn{1}{c}{} & - & \textbf{LR-MS:} 32 $ \times $ 32 $ \times $ 4 \\
         \multicolumn{1}{c}{} & \multicolumn{1}{c}{} & \multicolumn{1}{c}{} & \multicolumn{1}{c}{}& - & \textbf{Output:} 128 $ \times $ 128 $ \times $ 4 \\
         \hline

         \multicolumn{1}{c}{\multirow{8}{*}{QuickBird}} & \multicolumn{1}{c}{\multirow{8}{*}{11}} & \multicolumn{1}{c}{\multirow{8}{*}{3}} & \multicolumn{1}{c}{\multirow{4}{*}{Reduced}} & \textbf{\#Patches:} 1024 & \textbf{\#Patches:} 128 \\
         \multicolumn{1}{c}{} & \multicolumn{1}{c}{} & \multicolumn{1}{c}{} & \multicolumn{1}{c}{} & PAN: 128 $ \times $ 128 $ \times $ 1 & PAN: 128 $ \times $ 128 $ \times $ 1 \\
         \multicolumn{1}{c}{} & \multicolumn{1}{c}{} & \multicolumn{1}{c}{} & \multicolumn{1}{c}{} & \textbf{LR-MS:} 32 $ \times $ 32 $ \times $ 4 & \textbf{LR-MS:} 32 $ \times $ 32 $ \times $ 4 \\
         \multicolumn{1}{c}{} & \multicolumn{1}{c}{} & \multicolumn{1}{c}{} & \multicolumn{1}{c}{} & \textbf{Output:} 128 $ \times $ 128 $ \times $ 4 & \textbf{Output:} 128 $ \times $ 128 $ \times $ 4 \\
         \cline{4-6}
         \multicolumn{1}{c}{} & \multicolumn{1}{c}{} & \multicolumn{1}{c}{} & \multicolumn{1}{c}{\multirow{4}{*}{Full}} & - & \textbf{\#Patches:} 128 \\
         \multicolumn{1}{c}{} & \multicolumn{1}{c}{} & \multicolumn{1}{c}{} & \multicolumn{1}{c}{} & - & \textbf{PAN:} 128 $ \times $ 128 $ \times $ 1 \\
         \multicolumn{1}{c}{} & \multicolumn{1}{c}{} & \multicolumn{1}{c}{} & \multicolumn{1}{c}{} & - & \textbf{LR-MS:} 32 $ \times $ 32 $ \times $ 4 \\
         \multicolumn{1}{c}{} & \multicolumn{1}{c}{} & \multicolumn{1}{c}{} & \multicolumn{1}{c}{}& - & \textbf{Output:} 128 $ \times $ 128 $ \times $ 4 \\
         \hline

         \multicolumn{1}{c}{\multirow{8}{*}{GaoFen-2}} & \multicolumn{1}{c}{\multirow{8}{*}{11}} & \multicolumn{1}{c}{\multirow{8}{*}{1}} & \multicolumn{1}{c}{\multirow{4}{*}{Reduced}} & \textbf{\#Patches:} 1036 & \textbf{\#Patches:} 136 \\
         \multicolumn{1}{c}{} & \multicolumn{1}{c}{} & \multicolumn{1}{c}{} & \multicolumn{1}{c}{} & \textbf{PAN: } 128 $ \times $ 128 $ \times $ 1 & \textbf{PAN: } 128 $ \times $ 128 $ \times $ 1 \\
         \multicolumn{1}{c}{} & \multicolumn{1}{c}{} & \multicolumn{1}{c}{} & \multicolumn{1}{c}{} & \textbf{LR-MS:} 32 $ \times $ 32 $ \times $ 4 & \textbf{LR-MS:} 32 $ \times $ 32 $ \times $ 4 \\
         \multicolumn{1}{c}{} & \multicolumn{1}{c}{} & \multicolumn{1}{c}{} & \multicolumn{1}{c}{} & \textbf{Output:} 128 $ \times $ 128 $ \times $ 4 & \textbf{Output:} 128 $ \times $ 128 $ \times $ 4 \\
         \cline{4-6}
         \multicolumn{1}{c}{} & \multicolumn{1}{c}{} & \multicolumn{1}{c}{} & \multicolumn{1}{c}{\multirow{4}{*}{Full}} & -& \textbf{\#Patches:} 120 \\
         \multicolumn{1}{c}{} & \multicolumn{1}{c}{} & \multicolumn{1}{c}{} & \multicolumn{1}{c}{} & - & \textbf{PAN:} 128 $ \times $ 128 $ \times $ 1 \\
         \multicolumn{1}{c}{} & \multicolumn{1}{c}{} & \multicolumn{1}{c}{} & \multicolumn{1}{c}{} & - & \textbf{LR-MS: } 32 $ \times $ 32 $ \times $ 4 \\
         \multicolumn{1}{c}{} & \multicolumn{1}{c}{} & \multicolumn{1}{c}{} & \multicolumn{1}{c}{}& - & \textbf{Output:} 128 $ \times $ 128 $ \times $ 4 \\
         \hline
    \end{tabular}
    }
\label{tab:data}
\end{table*}

\begin{figure}[H]
    \centering
    \includegraphics[width=\linewidth]{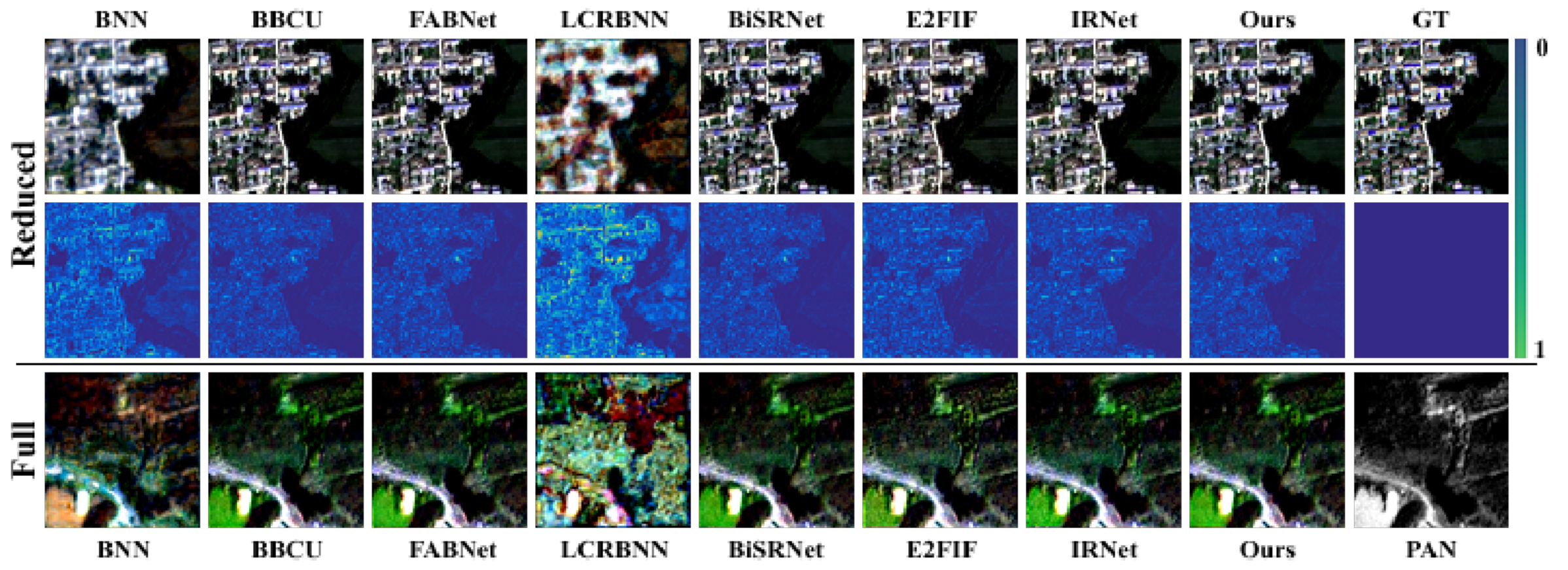}
    \caption{Visual comparison between our model and other binary methods on WV-2 example. The top two lines represent the reconstructed results and corresponding MAE maps of the reduced-resolution example, and the last line represents the reconstructed results of the full-resolution example.}
    \label{fig:wv2}
\end{figure}

\textbf{More experimental results.} To show the effectiveness of the proposed method, we also present the 
visualize the quantitative results and visual results of
the proposed method on other pan-sharpening datasets in Tab.~\ref{tab:qb}, Tab.~\ref{tab:wv2}, and Fig.~\ref{fig:wv2}.

\begin{table}
\caption{Quantitative comparison of our S2BNet with binary methods on the QuickBird dataset.}
\centering
\begin{tabular}{ccccccccc} 
\toprule
\multirow{2}{*}{Methods} & \multirow{2}{*}{Params (K)} & \multirow{2}{*}{FLOPs (G)} & \multicolumn{5}{c}{Reduced}                                                              & Full             \\
                         &                            &                          & PSNR$\uparrow$             & SSIM$\uparrow$            & $Q_4\uparrow$              & SAM$\downarrow$             & ERGAS$\downarrow$           & QNR$\uparrow$              \\ 
\hline
BNN                      & 17.760                     & 1.938                    & 23.4593          & 0.7868          & 0.4876          & 0.2729          & 7.3432          & 0.6854           \\
LCRBNN                   & 17.760                     & 1.938                    & 23.1537          & 0.7257          & 0.2900          & 0.2712          & 7.7950          & 0.5757           \\
BiSRNet                  & 26.459                     & 0.664                    & 30.6142          & 0.9252          & 0.8172          & \textbf{0.0602} & 3.6351          & 0.8468           \\
BBCU                     & 23.520                     & 0.664                    & 31.1226          & 0.9340          & 0.8398          & 0.0846          & 3.7487          & 0.8519           \\
FABNet                   & 59.296                     & 0.664                    & 32.3291          & 0.9338          & 0.8361          & 0.0625          & 3.0356          & \textbf{0.8555}  \\
\textbf{S2BNet}                   & 75.407                     & 0.920                    & \textbf{32.7623} & \textbf{0.9340} & \textbf{0.8411} & 0.0675          & \textbf{2.9236} & 0.8495           \\
\bottomrule
\end{tabular}
\label{tab:qb}
\end{table}

\begin{table}
\caption{Quantitative comparison of our S2BNet with our full-precision and binary methods on the WorldView-2 dataset.}
\centering
\setlength{\extrarowheight}{0pt}
\addtolength{\extrarowheight}{\aboverulesep}
\addtolength{\extrarowheight}{\belowrulesep}
\setlength{\aboverulesep}{0pt}
\setlength{\belowrulesep}{0pt}
\resizebox{\linewidth}{!}{
\begin{tabular}{c|c|cc|ccccc|ccc} 
\toprule
\rowcolor[rgb]{0.922,0.922,0.922} {\cellcolor[rgb]{0.922,0.922,0.922}}                           & {\cellcolor[rgb]{0.922,0.922,0.922}}                         & {\cellcolor[rgb]{0.922,0.922,0.922}}                            & {\cellcolor[rgb]{0.922,0.922,0.922}}                          & \multicolumn{5}{c|}{Reduced-Resolution}                                                  & \multicolumn{3}{c}{Full-Resolution}                  \\
\rowcolor[rgb]{0.922,0.922,0.922} \multirow{-2}{*}{{\cellcolor[rgb]{0.922,0.922,0.922}}Category} & \multirow{-2}{*}{{\cellcolor[rgb]{0.922,0.922,0.922}}Method} & \multirow{-2}{*}{{\cellcolor[rgb]{0.922,0.922,0.922}}Params (K)} & \multirow{-2}{*}{{\cellcolor[rgb]{0.922,0.922,0.922}}FLOPs (G)} & PSNR$\uparrow$             & SSIM$\uparrow$             & $Q_4\uparrow$                & SAM$\downarrow$          & ERGAS$\downarrow$           & $D_\lambda\downarrow$        & $D_s\downarrow$            & QNR$\uparrow$              \\ 
\hline
\rowcolor[rgb]{0.941,1,0.941} {\cellcolor[rgb]{0.941,1,0.941}}                                   & DMFNet ~\cite{Yuan2019DMFNet}                                                      & 1631.924                                                        & 145.811                                                       & 41.0268          & 0.9716          & 0.8184          & 0.0252          & 1.0855          & 0.0644          & 0.0863          & 0.8557           \\
\rowcolor[rgb]{0.941,1,0.941} {\cellcolor[rgb]{0.941,1,0.941}}                                   & PanFormer ~\cite{zhou2022panformer}                                                    & 1530.300                                                          & 12.002                                                        & 41.3495          & 0.9731          & 0.8237          & 0.0242          & 1.0621          & 0.0628          & 0.0844          & 0.8590           \\
\rowcolor[rgb]{0.941,1,0.941} {\cellcolor[rgb]{0.941,1,0.941}}                                   & MutInf ~\cite{zhou2022spatial}                                                       & 185.496                                                         & 9.986                                                         & 41.9527          & 0.9760          & 0.8258          & 0.0227          & 1.0152          & 0.0622          & 0.0794          & 0.8643           \\
\rowcolor[rgb]{0.941,1,0.941} {\cellcolor[rgb]{0.941,1,0.941}}                                   & FameNet ~\cite{DBLP:conf/aaai/HeYLX0Z24}                                                     & 1244.228                                                        & 39.272                                                        & 42.0278          & 0.9768          & 0.8332          & 0.0222          & 0.9936          & 0.0624          & 0.0753          & 0.8678           \\
\rowcolor[rgb]{0.941,1,0.941} {\cellcolor[rgb]{0.941,1,0.941}}                                   & CANNet ~\cite{Duan2024CANNet}                                                      & 785.118                                                         & 3.237                                                         & 41.5868          & 0.9737          & 0.8256          & 0.0237          & 1.0517          & 0.0612          & 0.0767          & 0.8679           \\
\rowcolor[rgb]{0.941,1,0.941} {\cellcolor[rgb]{0.941,1,0.941}}                                   & UGCC ~\cite{Zeng2025Cross-Modal}                                                        & 233.233                                                         & 307.495                                                       & 35.8744          & 0.9296          & 0.6827          & 0.0460          & 1.9253          & 0.0887          & 0.0872          & 0.8327           \\
\rowcolor[rgb]{0.941,1,0.941} \multirow{-7}{*}{{\cellcolor[rgb]{0.941,1,0.941}}Full}            & MSCSCFormer ~\cite{Ye2024MSCSCformer}                                                 & 1950.686                                                        & 582.435                                                       & 40.5338          & 0.9691          & 0.7963          & 0.0270          & 1.2016          & 0.0654          & 0.0772          & 0.8634           \\ 
\hline
\rowcolor[rgb]{1,0.969,0.969} {\cellcolor[rgb]{1,0.969,0.969}}                             & BNN ~\cite{Hubara2016BNN}                                                         & 17.760                                                          & 1.938                                                         & 32.9122          & 0.8706          & 0.4595          & 0.0679          & 3.1063          & 0.0970          & 0.1244          & 0.7916           \\
\rowcolor[rgb]{1,0.969,0.969} {\cellcolor[rgb]{1,0.969,0.969}}                             & LCRBNN ~\cite{shang2022lipschitz}                                                      & 17.760                                                          & 1.938                                                         & 29.3907          & 0.7781          & 0.1877          & 0.0886          & 4.9267          & 0.0968          & 0.2997          & 0.6323           \\
\rowcolor[rgb]{1,0.969,0.969} {\cellcolor[rgb]{1,0.969,0.969}}                             & BiSRNet ~\cite{cai2023binarized}                                                     & 26.459                                                          & 0.664                                                         & 40.5263          & 0.9680          & 0.8016          & 0.0270          & 1.1755          & 0.0631          & 0.0756          & 0.8670           \\
\rowcolor[rgb]{1,0.969,0.969} {\cellcolor[rgb]{1,0.969,0.969}}                             & BBCU ~\cite{Xia2022BBCU}                                                        & 23.520                                                          & 0.664                                                         & 40.4211          & 0.9675          & 0.7957          & 0.0272          & 1.1904          & 0.0627          & 0.0758          & 0.8671           \\
\rowcolor[rgb]{1,0.969,0.969} {\cellcolor[rgb]{1,0.969,0.969}}                             & FABNet ~\cite{Jiang2023FABNet}                                                      & 59.296                                                          & 0.664                                                         & 40.0005          & 0.9647          & 0.7898          & 0.0287          & 1.2297          & 0.0624          & \textbf{0.0748} & \textbf{0.8684}  \\
\rowcolor[rgb]{1,0.969,0.969} {\cellcolor[rgb]{1,0.969,0.969}}                             & IRNet ~\cite{Qin2020IRNet}                                                       & 17.760                                                          & 1.938                                                         & 39.2731          & 0.9615          & 0.7736          & 0.0311          & 1.3434          & 0.0687          & 0.0801          & 0.8577           \\
\rowcolor[rgb]{1,0.969,0.969} {\cellcolor[rgb]{1,0.969,0.969}}                             & E2FIF ~\cite{Song2023E2FIF}                                                       & 20.640                                                          & 0.664                                                         & 37.1290          & 0.9407          & 0.7273          & 0.0412          & 1.6997          & 0.0741          & 0.0833          & 0.8497           \\
\rowcolor[rgb]{1,0.969,0.969} \multirow{-8}{*}{{\cellcolor[rgb]{1,0.969,0.969}}Binary}     & \textbf{S2BNet}                                                          & 75.407                                                          & 0.920                                                         & \textbf{40.6059} & \textbf{0.9684} & \textbf{0.8035} & \textbf{0.0267} & \textbf{1.1742} & \textbf{0.0619} & 0.0760          & 0.8677           \\
\bottomrule
\end{tabular}
}
\label{tab:wv2}
\end{table}

\end{document}